\pdfoutput=1
\documentclass[runningheads]{llncs}
\usepackage[T1]{fontenc}
\usepackage{graphicx}
\usepackage{booktabs}
\usepackage{amsmath}
\usepackage[hidelinks]{hyperref}
\usepackage{longtable,booktabs,array}
\usepackage{calc}
\usepackage{amssymb}
\usepackage{adjustbox}

\usepackage{newunicodechar}
\newunicodechar{−}{\ensuremath{-}}
\newunicodechar{→}{\ensuremath{\rightarrow}}
\newunicodechar{§}{\S{}}
\newunicodechar{×}{\ensuremath{\times}}
\newunicodechar{Δ}{\ensuremath{\Delta}}
\newunicodechar{≈}{\ensuremath{\approx}}
\newunicodechar{ρ}{\ensuremath{\rho}}
\newunicodechar{≠}{\ensuremath{\neq}}
\newunicodechar{⇒}{\ensuremath{\Rightarrow}}
\newunicodechar{‖}{\ensuremath{\|}}
\newunicodechar{∈}{\ensuremath{\in}}
\newunicodechar{≳}{\ensuremath{\gtrsim}}
\newunicodechar{≫}{\ensuremath{\gg}}
\newunicodechar{↓}{\ensuremath{\downarrow}}
\newunicodechar{≥}{\ensuremath{\geq}}
\newunicodechar{≤}{\ensuremath{\leq}}
\usepackage{placeins} 
\begin{document}
\title{Can You Trust Frozen Hematology\\
Foundation Models\\
under Acquisition Shift?\thanks{Accepted at the HemaRAI 2026 workshop (a MICCAI 2026 satellite event), oral presentation; to appear in \emph{MICCAI 2026 Satellite Events}, Lecture Notes in Computer Science, Springer. Project page: \url{https://jaishrm07.github.io/hematology-fm-robustness/}}}
\titlerunning{Can You Trust Hematology Foundation Models under Acquisition Shift?}
\author{Jai Kumar Sharma\inst{1} \and Peeyush Tapadiya\inst{2}}
\authorrunning{J.\,K. Sharma and P. Tapadiya}
\institute{Virginia Tech \\ \email{jaisharma@vt.edu}
\and Accenture \\ \email{peeyush.tapadiya@accenture.com}}
\maketitle

\begin{abstract}
Frozen hematology foundation-model (FM) embeddings reach near-saturated in-domain white-blood-cell (WBC) accuracy, but
clinical deployment demands reliability across scanners, sites, stains and preparation pipelines. We audit 15 frozen
encoders (hematology, pathology, and general vision encoders) across four public single-cell acquisition domains
along two axes: accuracy robustness and calibration. In-domain linear-probe macro-F1 is saturated (0.98--0.997), yet
cross-dataset macro-F1 drops 34--72\% and rankings re-order: DinoBloom-L, the in-domain best, falls to 10th of 15 on the
most-shifted target (MLL23) at the benchmark's shared 224-px input, while RedDino and several general and pathology encoders outrank it. Rank transfer is
\emph{probe-dependent}: 1-NN retrieval is more stable on average than a source-fitted linear head (median $\rho$ 0.65 vs
0.45), but neither clean-domain probe universally predicts target robustness. Calibration also collapses:
source-trained probes are nearly calibrated in-domain (Expected Calibration Error [ECE] 0.004) but become confidently wrong off-domain (ECE 0.35), and
source-fitted temperature scaling transfers poorly. We further audit pretraining exposure and identify MLL23 as
corresponding to DinoBloom's internal cohort; because the only DinoBloom-held-out dataset is also our source domain, this
benchmark cannot isolate exposure from scanner-associated distribution shift. Finally, label-free adaptation and marginal-entropy-based model selection appear safe under balanced
evaluation but fail under realistic WBC class-prior shift. \emph{Class-Balanced Re-standardization} (CBR), a training-free
pseudo-label-balanced feature normalization, improves all evaluated target-prior scenario means and partially improves
calibration, although encoder-level exceptions and residual miscalibration remain. These results argue that hematology FM benchmarks must jointly
audit accuracy, calibration, exposure, and class-prior robustness.
\end{abstract}

\section{Introduction}
Blood-smear classifiers are typically trained and benchmarked on one acquisition pipeline but deployed across scanners,
laboratories, stains, and preparation protocols. For white-blood-cell (WBC) differential support this creates a dangerous
failure mode: a model can be accurate on its development scanner yet systematically (and confidently) wrong
elsewhere, where a second site cannot sanity-check it~\cite{tsutsui,scangen}.

Hematology foundation models (FMs) are commonly compared using frozen linear-probe or nearest-neighbor accuracy on
in-domain splits~\cite{dinobloom,reddino}. Such evaluation does not answer the questions that matter for deployment: whether clean rankings survive
real acquisition-domain shift, whether predicted confidence remains calibrated off-domain, whether benchmark targets
overlap the model's pretraining, or whether label-free test-time adaptation stays safe under the class imbalance of a
real WBC differential.

We therefore \emph{audit} frozen embeddings rather than train new classifiers. We train a source-domain linear probe (and
a 1-NN probe), evaluate zero-shot across four public single-cell acquisition domains, and measure reliability along two
axes: accuracy and calibration. We additionally audit dataset \emph{exposure} and stress-test label-free adaptation
under balanced and realistic target class priors.

\noindent\textbf{Contributions.} (1) A cross-domain \emph{accuracy} audit of 15 frozen encoders (hematology FMs, pathology
FMs, general SSL/VLM/ImageNet), showing that saturated in-domain linear-probe accuracy does not identify the robust
encoder under scanner-associated shift, and that rank transfer is \emph{probe-dependent} (1-NN ranks are more stable on
average than a linear head, yet neither probe is a universally reliable clean-domain selector). (2) A \emph{calibration} audit showing
source-trained probes become confidently wrong off-domain and that source-fitted temperature scaling transfers poorly.
(3) A \emph{pretraining-exposure} audit showing public hematology benchmarks are exposure-ambiguous (we identify MLL23 as
corresponding to DinoBloom's internal cohort), finding that these four datasets cannot isolate exposure from
scanner-associated shift. (4) A label-shift stress test showing that standard label-free adaptation
and marginal-entropy-based model selection, although effective under balanced evaluation, fail under realistic WBC priors. We evaluate
\textbf{Class-Balanced Re-standardization (CBR)}, a training-free, frozen-feature instance of class-balanced
normalization~\cite{balancedbn,csnorm}, which improves all evaluated target-prior scenario means and partially improves
calibration.

\section{Related work}
\textbf{Hematology FMs and WBC robustness.} DinoBloom~\cite{dinobloom} and the RBC-specialized RedDino~\cite{reddino} are
recent, widely used frozen hematology encoders; Tsutsui et al.~\cite{tsutsui} show supervised WBC CNNs degrade across imaging
conditions, and WBCBench~\cite{wbcbench} benchmarks robust WBC classification under \emph{class imbalance}. We instead
audit \emph{frozen} embeddings under \emph{real} cross-dataset acquisition shift and stress-test class-prior robustness.

\textbf{Scanner/site robustness in pathology FMs.} ScanGen~\cite{scangen} and scanner-induced-shift work~\cite{scannershift}
establish scanner sensitivity for \emph{tissue} FMs, attributing it to embedding/calibration drift rather than leakage.
These motivate scanner/site reliability evaluation but do not study frozen single-cell hematology embeddings, calibration
transfer, exposure ambiguity, or class-prior stress tests.

\textbf{Calibration and exposure audits.} Temperature scaling~\cite{guo} and the off-distribution degradation of
calibration~\cite{ovadia} are established in general; we also report Expected Calibration Error (ECE), adaptive-ECE~\cite{nixon}, NLL and Brier (Sec.~5),
port these to frozen hematology FMs, and add the \emph{transfer} test: does a source-fitted temperature survive a
scanner change? PathBench~\cite{pathbench} \emph{prevents}
pretraining-data leakage by strict eval/pretraining separation; we instead \emph{audit} pretraining overlap and report
it. Accuracy-on-the-Line~\cite{accline} holds in-distribution, while Accuracy-on-the-\emph{Wrong}-Line~\cite{wrongline}
shows it can break via label noise/nuisance features; we add a hematology cross-dataset case. (Effective
robustness~\cite{taori} and ImageNet-C~\cite{imagenetc} are the natural-image precedents.)

\textbf{Label-free TTA and model selection.} Our adaptation baseline is feature-space BN/AdaBN-style
adaptation~\cite{adabn,bnadapt} (cf.\ CORAL~\cite{coral}, Tent~\cite{tent}, SHOT~\cite{shot}, and class-aware feature
alignment~\cite{cafa}); \emph{class-balanced / label-shift-robust normalization}~\cite{balancedbn,csnorm} is an
established sub-line we build on directly. We claim no new principle: CBR is its simplest frozen-FM, single-batch,
training-free instance; unlike CAFA~\cite{cafa} it updates no weights, only re-estimating frozen statistics in a
pseudo-label-balanced way. Our contributions are the first-moment
diagnosis and the demonstration that vanilla BN-adaptation, SHOT/IM, and the label-shift method BBSE all silently fail
under realistic class imbalance on hematology FMs, while the class-balanced instance does not. We also test a label-free
encoder-\emph{selection} statistic (marginal entropy), a \emph{cautionary negative}, and compare with
agreement-on-the-line performance prediction~\cite{agreementline,aglfm} (Sec.~5).

\section{Method}
\textbf{Protocol.} We freeze each encoder, extract its CLS/pooled features, fit source-domain standardization, and train an
$L_2$-regularized logistic probe (and a 1-NN probe) on a source domain, then evaluate zero-shot transfer to held-out
target acquisition domains. All encoders run at $224\times224$ (DinoBloom forced to 224, overriding the checkpoint's 518
default; resolution sensitivity in Suppl.\ A.14). We use the fixed five-class WBC intersection and evaluate the full
source$\times$target matrix, including a reverse-direction control.

\textbf{Metrics.} Our primary metric is macro-F1. We additionally report the Spearman rank correlation between in-domain
and target encoder rankings, effective robustness, relative gap, per-class recall, and bootstrap 95\% CIs over five
source splits.

\textbf{Label-shift evaluation.} Because real blood differentials are class-imbalanced, we evaluate all label-free
methods on the balanced target \emph{and} on targets resampled to realistic/skewed class priors. The exact clinical and
neutrophil-heavy priors are listed in Suppl.\ A.12.3; they mimic peripheral-blood differentials and stress
neutrophil-dominated deployment batches.

\textbf{Test-time adaptation.} Given a probe trained in source-standardized space, a standard label-free adaptation
re-standardizes \emph{target} embeddings with their own \emph{unlabeled} batch statistics (\texttt{tgtstd}; feature
BN-adaptation). Because the batch mean/std are dominated by the majority class, this breaks under label shift (Sec.~5). Our
\textbf{CBR} estimates target statistics from \emph{pseudo-label-balanced} class means $\bar\mu{=}\frac1{|\mathcal C|}\sum_c
\mathrm{mean}(X_t[\hat c{=}c])$ and pooled within-class std $\bar\sigma$ (equal-weighting predicted classes
\emph{reduces} dependence on the target class prior, conditional on pseudo-label quality and class coverage; pseudo-labels
$\hat c$ from the source probe), then transforms $(X_t-\bar\mu)/\bar\sigma$, still label-free, training-free,
single-batch. Empty pseudo-classes are omitted; singleton pseudo-classes contribute to the mean but not the pooled
variance; if all target cells collapse to one pseudo-class we fall back to the source standard deviation.

\section{Experimental setup}
Source: Acevedo (PBC)~\cite{acevedo} via BloodMNIST@224~\cite{medmnist} (CellaVision DM96), 10{,}298 WBC images. Targets
(zero-shot): MLL23/Metafer~\cite{mll23}, Matek-LMU/M8~\cite{matek}, Raabin~\cite{raabin} (smartphone+Olympus).
\textbf{Encoders} (15 frozen; families and full names in Table~\ref{tab:leaderboard}): DinoBloom~\cite{dinobloom},
RedDino~\cite{reddino}, Phikon~\cite{phikon}, Lunit-DINO~\cite{lunit}, DINOv2~\cite{dinov2}, EVA-02~\cite{eva02},
BiomedCLIP~\cite{biomedclip}, CLIP~\cite{clip}, ImageNet ViT-B~\cite{vit} and ResNet-50~\cite{resnet}; plus a supervised ResNet-18 baseline.
\textbf{Pretraining-exposure audit:} DinoBloom reports training on all
datasets \emph{except} Acevedo; its internal ``LabAnonymous'' cohort is MLL23~\cite{dinobloom,mll23} (same 41{,}906-image Munich Leukemia
Laboratory dataset; the MLL23 descriptor cites DinoBloom as a prior user). So for DinoBloom, Acevedo (our source) is the
only held-out dataset and Matek, Raabin and MLL23 are all in-pretraining, so there is no leakage-free \emph{target}; we audit
exposure rather than assume it. For the non-hematology encoders we found no documented \emph{dedicated} exposure to these
blood-cell datasets, though web-scale pretraining (e.g.\ CLIP) cannot be fully ruled out. We split at the image level
(patient identifiers are not consistently available across these public datasets): the probe is trained on a 70\%
stratified split and the in-domain test is the held-out 30\%, used only as a clean-ranking proxy; all deployment claims
rest on cross-dataset transfer.

\section{Results}
\begin{table}[t]
\centering
\caption{Cross-dataset leaderboard (linear-probe macro-F1, 5 seeds; source = Acevedo), sorted by
MLL23. The ``MLL23 \#'' column makes the re-ranking legible: the in-domain \#1 (DinoBloom-L)
falls to 10th of 15 on the most-shifted target ($0.15$ macro-F1 behind the best encoder, RedDino). Bold = strict column max.
The supervised ResNet-18 (scratch, below the rule) is a non-frozen baseline, not part of the
15-encoder ranking. Paired-bootstrap 95\% CIs for the key MLL23 comparisons (Suppl.\ A.9) all exclude zero.
Exposure status audited in Sec.~4.}
\label{tab:leaderboard}
{\small\setlength{\tabcolsep}{4pt}
\begin{tabular}{lccccc}
\toprule
encoder & Acevedo$^\ddagger$ & Matek & MLL23 & MLL23\,\# & Raabin \\
\midrule
RedDino \textit{(hema)} & 0.994 & 0.544 & \textbf{0.704} & 1 & \textbf{0.450} \\
DinoBloom-S \textit{(hema)}  & 0.995 & 0.613 & 0.671 & 2 & 0.341 \\
DINOv2-B           & 0.991 & 0.595 & 0.650 & 3 & 0.444 \\
DINOv2-S           & 0.988 & 0.505 & 0.634 & 4 & 0.248 \\
Lunit-DINO \textit{(path)} & 0.995 & 0.409 & 0.609 & 5 & 0.288 \\
DINOv2-L           & 0.991 & 0.555 & 0.588 & 6 & 0.315 \\
ViT-B \textit{(IN)}   & 0.993 & 0.616 & 0.571 & 7 & 0.439 \\
CLIP-L/14          & 0.986 & 0.557 & 0.568 & 8 & 0.290 \\
DinoBloom-B \textit{(hema)} & \textbf{0.997} & 0.635 & 0.553 & 9 & 0.448 \\
DinoBloom-L \textit{(hema)}$^\dagger$ & \textbf{0.997} & \textbf{0.648} & 0.552 & 10 & 0.385 \\
BiomedCLIP         & 0.980 & 0.410 & 0.526 & 11 & 0.280 \\
EVA-02 \textit{(IN)}  & 0.991 & 0.486 & 0.486 & 12 & 0.378 \\
ResNet-50 \textit{(IN)} & 0.980 & 0.357 & 0.416 & 13 & 0.304 \\
Phikon \textit{(path)} & 0.995 & 0.448 & 0.410 & 14 & 0.265 \\
CLIP-B/16          & 0.981 & 0.387 & 0.386 & 15 & 0.337 \\
\midrule
\textit{Sup.\ ResNet-18 (scratch)} & \textit{0.908} & \textit{0.584} & \textit{0.255} & \textit{--} & \textit{0.096} \\
\bottomrule
\end{tabular}}\\[2pt]
{\footnotesize \textit{hema}=hematology FM, \textit{path}=pathology FM, \textit{IN}=ImageNet-supervised.
$^\ddagger$Acevedo is the in-domain source; $^\dagger$in-domain \#1 by unrounded macro-F1 (rounds to a tie with
DinoBloom-B). ``rank'' = position on the most-shifted target MLL23 (1 = best of 15).}
\end{table}

\begin{figure}[t]\centering
\includegraphics[width=.78\textwidth]{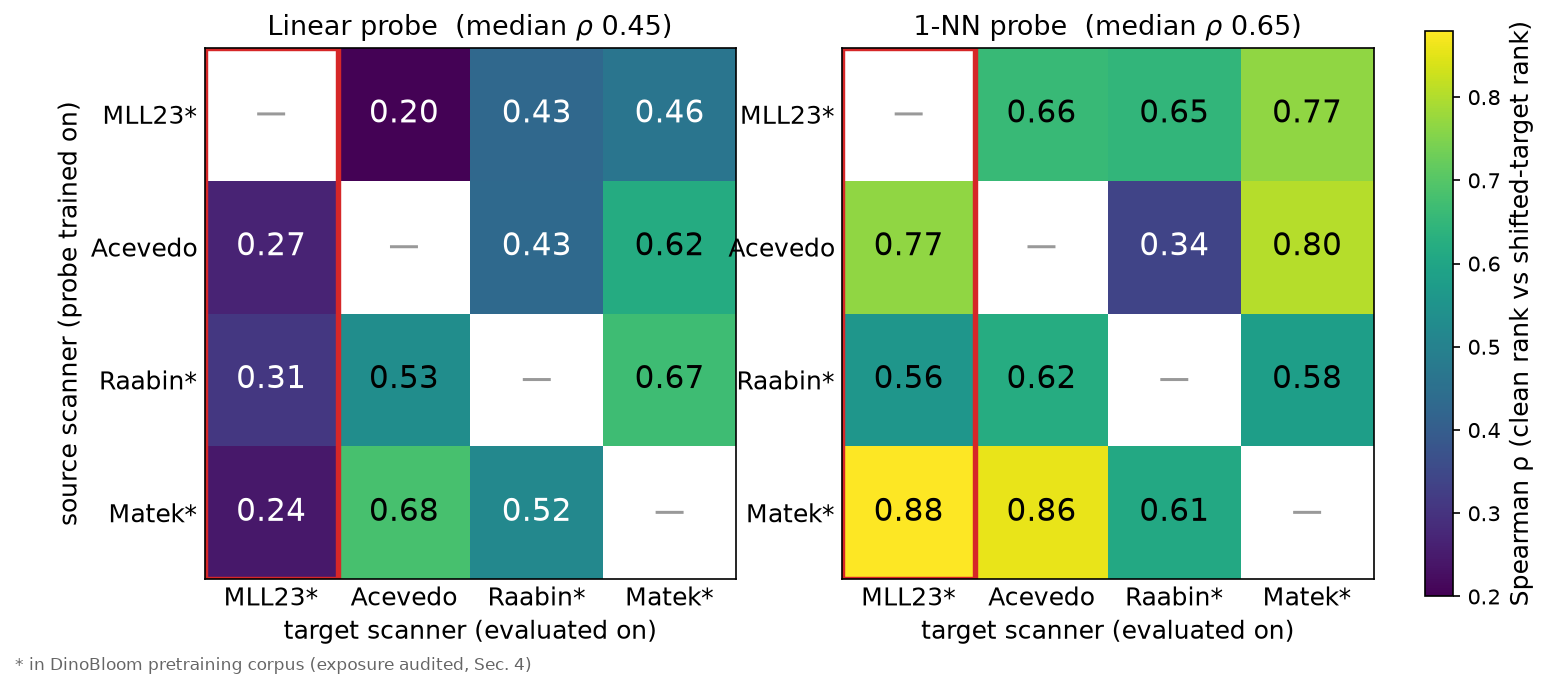}
\caption{Clean-to-target rank transfer is \emph{probe-dependent}. Spearman $\rho$ between in-domain and shifted-target
encoder rankings ($n{=}15$), linear (left) and 1-NN (right). 1-NN is more stable on average (median $\rho$ 0.65 vs 0.45)
but not uniformly predictive (Acevedo$\to$Raabin 0.34). MLL23 (red, largest shift) is hardest for both and is \emph{not}
leakage-free for DinoBloom (Sec.~4; ${*}$ = in DinoBloom pretraining corpus).}\label{fig:rho}
\end{figure}

\textbf{Axis A: clean accuracy mis-ranks encoders.} In-domain macro-F1 is saturated (0.98--0.997) yet cross-dataset macro-F1
drops 34--72\% and re-orders: DinoBloom-L (in-domain \#1) falls to 10/15 on MLL23 at the shared 224-px input (mid-rank
at native 518, still below DinoBloom-S and RedDino; Suppl.\ A.14), beaten by RedDino (0.704) and even a pathology FM;
the two pathology FMs diverge sharply (Lunit 0.609 vs Phikon 0.410), so specialization alone does not predict
robustness. These gaps are statistically stable: paired bootstraps over the MLL23
examples put the RedDino$-$DinoBloom-L, DinoBloom-S$-$DinoBloom-L, and Lunit$-$Phikon differences at 95\% CIs that
exclude zero (Suppl.\ A.9). Spearman $\rho$(in-domain, cross-dataset rank) is as low as 0.27 on the
most-shifted target (Fig.~\ref{fig:rho}), so clean accuracy cannot discriminate robustness. This is not specific to the Acevedo
source: across the full source$\times$target matrix (Suppl.\ A.15), $\rho$ stays low (median 0.45) and the in-domain-best encoder
is dethroned in 8 of 12 source--target pairs. \emph{Rank transfer is probe-dependent}: across the same 12 pairs 1-NN is
more stable on average (median $\rho$ 0.65 vs the linear 0.45; Suppl.\ A.15), so local retrieval geometry transfers more
consistently, but 1-NN is not universal either ($\rho$ 0.34 on Acevedo$\to$Raabin) and is not a reliable clean-domain
selector. This holds across logistic-$C$ and linear-SVM heads (Suppl.\ A.16).

\FloatBarrier
\textbf{Exposure audit: no clear leakage-inflation signature.} Pretraining exposure is an obvious confound, but the
available datasets do not identify its effect cleanly. All three transfer targets were used in DinoBloom training
(Sec.~4), so there is no leakage-free DinoBloom target: its target-side numbers measure transfer to
\emph{in-pretraining} domains, not clean held-out generalization. Conversely, in the reverse-source matrix (Suppl.\ A.15) DinoBloom
transfers \emph{best} to its held-out Acevedo and is not dominant on its training targets; this is inconsistent with a
simple leakage-inflation explanation, but Acevedo is also an easier source/target domain, so exposure and domain
difficulty remain confounded. A controlled fine-tuning experiment (Suppl.\ A.5) confirms scanner exposure can
selectively improve same-domain transfer. We therefore report exposure status without claiming or excluding a
DinoBloom-specific leakage effect.

\begin{figure}[t]\centering
\includegraphics[width=.92\textwidth]{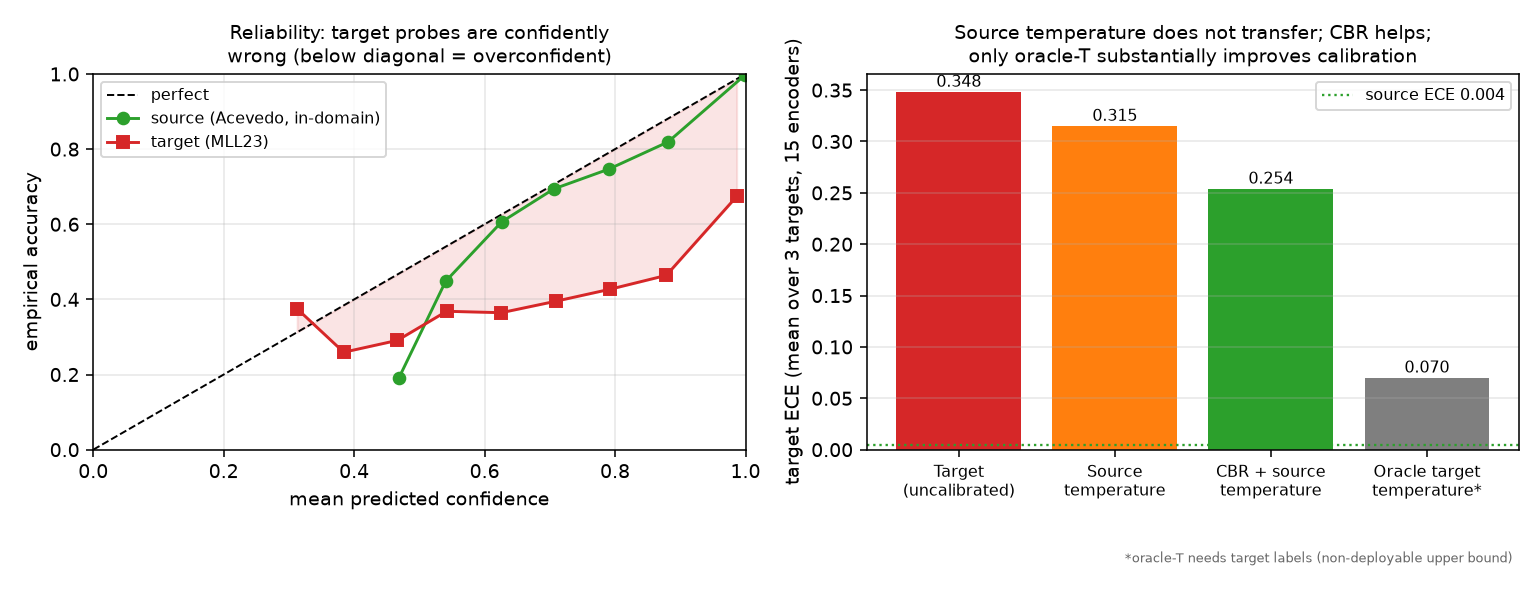}
\caption{Axis B (calibration). Left: reliability diagram (15 encoders pooled; Acevedo$\to$MLL23): target probes sit
below the diagonal (confidently wrong). Right: target ECE by arm (mean over 3 targets $\times$ 15 encoders): source
temperature scaling barely helps; CBR\,+\,source-T partially helps; only oracle target-T (needs target labels, not
deployable) substantially improves it, still above in-domain.}\label{fig:cal}
\end{figure}

\textbf{Axis B: calibration collapses and source calibration does not transfer (Fig.~\ref{fig:cal}).} Source-trained
probes are near-perfectly calibrated in-domain (ECE 0.004, NLL 0.03) but collapse off-domain (mean over 15 encoders
$\times$ 3 targets: ECE 0.35, NLL 3.2; full metrics in Suppl.\ A.13). Crucially, \textbf{a temperature
fitted on the held-out source does not transfer} to the target (ECE $0.35{\to}0.32$), since the source probe is already
calibrated (T$\approx$1); only \emph{oracle} target temperature scaling
\emph{substantially improves} it (ECE$\to$0.07), so calibration must be corrected per scanner. CBR alone improves target
ECE 0.35$\to$0.29, and adding the source-fitted temperature after CBR lowers it to 0.25 (43/45 encoder$\times$target
cells improve; Suppl.\ A.13), though substantial
miscalibration remains ($\sim$60$\times$ in-domain); CBR thus mitigates but does not solve calibration, and the two axes
are distinct deployment problems.

\textbf{Mechanism.} Is the failure explained by synthetic stain/color perturbation, or by a measurable embedding
shift? Synthetic stain corruptions do not reproduce the real failure (DinoBloom is
stain-corruption-invariant yet collapses on Metafer); the dominant measurable effect is a shift in the per-feature means
(the embedding's first moment). Per-class, DinoBloom-B/L collapse on lymphocytes off-distribution
(recall 0.16--0.20, mis-called neutrophils) while DINOv2-B retains them (recall 0.82, 5-seed mean).

\textbf{Label-free test-time adaptation: balanced evaluation hides failure under clinical priors.} Because the measurable
shift is low-order, we test whether label-free feature-statistics adaptation can mitigate it, in one canonical experiment
(18 target$\times$prior scenarios; Suppl.\ A.12). On balanced and mild priors
global target standardization (\texttt{tgtstd}) helps, but on realistic neutrophil-dominated priors it \emph{hurts}
(clinical $-0.07$, neutrophil-heavy $-0.08$; hurts in 6/18 scenarios). Learned SHOT/IM is worse (mean $-0.029$, hurts
10/18), and a standard label-shift estimator, BBSE~\cite{bbse}, hurts in \emph{all} 18 (mean $-0.035$): it corrects the
\emph{label prior}, not the feature-space scanner shift. The failure is \emph{not a pure label-prior shift solvable in
prediction space}; a \emph{class-balanced feature} statistic is better matched to it.

The class-prior problem also affects model selection: a label-free \emph{selection} heuristic (marginal-prediction
entropy) appears oracle-like under balanced target sampling but incurs 0.25--0.37 selection regret under skewed priors
(Suppl.\ A.10--A.11): we identify no reliable label-free deployment selector.

\begin{figure}[t]\centering
\includegraphics[width=.62\textwidth]{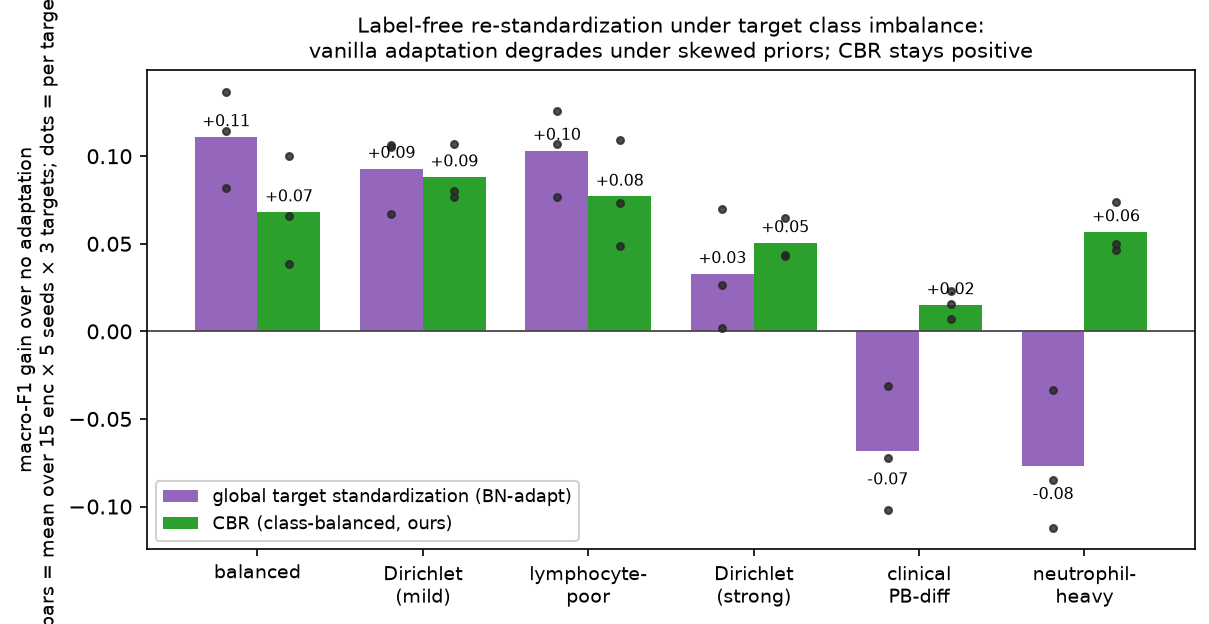}
\caption{Label-free re-standardization under target class imbalance, from one canonical experiment (15 encoders $\times$
5 source seeds $\times$ 25 draws $\times$ 3 targets $\times$ 6 priors). Global target standardization (BN-adaptation)
hurts on realistic neutrophil-dominated priors, while CBR (class-balanced, ours) is positive in all 18 \emph{evaluated}
target-prior scenario means (bars = mean gain over no adaptation, dots = per target; per-encoder breakdown:
Suppl.\ A.12.1).}\label{fig:cbr}
\end{figure}

For adaptation, however, class-balanced target statistics help: \textbf{CBR} (which equal-weights
predicted classes to \emph{reduce} dependence on the target class prior) is positive in all 18 evaluated
target$\times$prior scenarios (mean $+0.059$, range $[+0.007,+0.109]$, hurt 0/18; hierarchical bootstrap CI
$[+0.046,+0.073]$, Suppl.\ A.12.6; Fig.~\ref{fig:cbr}). These are scenario means over 15 encoders: per encoder, mean
gain is positive for all 15, though 29/270 encoder$\times$scenario cells are negative (worst $-0.09$, mostly DinoBloom
under extreme skew; vs 96/270 for \texttt{tgtstd}, 149/270 for IM; Suppl.\ A.12.1).
An ablation shows the balanced first-moment term drives the gain
(CBR-mean $+0.053$); CBR recovers 61\% of the true-label oracle and CBR$+$BBSE $\approx$ CBR. Small-batch stress tests
(Suppl.\ A.12.5) show CBR stays positive on average even at $K{=}16$ but is noisy for tiny skewed batches; we recommend
$K\geq32$.

\section{Discussion and limitations}
\textbf{Deployment implications.} Select encoders using representative cross-domain validation rather than in-domain
accuracy; disclose pretraining overlap; recalibrate confidence per scanner/site; and prefer class-balanced over global
target-statistics adaptation when target batches are imbalanced.
\textbf{Limitations.} The five-class WBC intersection; modest power for individual $\rho$ over 15 encoders (we rely on
CI-separated spread and rank re-ordering, and describe \emph{scanner-associated cross-dataset} shift); no leakage-free
DinoBloom target; probe-dependent selection failure (1-NN more stable on average but not universally reliable);
transductive, pseudo-label-dependent CBR ($\sim$61\% of the oracle, $K\geq32$, with per-encoder exceptions under extreme
skew, Suppl.\ A.12.1); and one stain family per dataset.

\section{Conclusion}
In-domain linear-probe accuracy does not predict which frozen encoder is robust under scanner-associated shift; rank
transfer is probe-dependent, both target accuracy and calibration degrade, and benchmarks are exposure-ambiguous. Class-balanced
re-standardization gives a partial mitigation; we urge jointly auditing accuracy, calibration, exposure, and class-prior
robustness for hematology FMs.

\subsubsection*{\discintname}
The authors have no competing interests to declare that are relevant to the content of this article.

\clearpage
\setcounter{figure}{0}\setcounter{table}{0}
\renewcommand{\thefigure}{S\arabic{figure}}\renewcommand{\thetable}{S\arabic{table}}
\phantomsection\pdfbookmark[1]{Supplementary Material}{supp}
\section*{Supplementary Material}

This supplement provides supporting analyses for the main paper. Section
labels A.1--A.16 match the citations in the main text. Unless noted,
experiments follow the main-paper protocol: frozen embeddings, Acevedo
as source, 5 source splits, and the 15-encoder benchmark. Mechanism
probes in A.1--A.6 use the core 11-encoder set; A.7--A.16 use the full
15-encoder benchmark. The main paper is self-contained; this document
provides additional controls, stress tests, and implementation details.

\begin{table}[ht]\centering\footnotesize\begin{adjustbox}{max width=\textwidth}\begin{tabular}{@{}
  >{\raggedright\arraybackslash}p{(\columnwidth - 4\tabcolsep) * \real{0.3333}}
  >{\raggedright\arraybackslash}p{(\columnwidth - 4\tabcolsep) * \real{0.3333}}
  >{\raggedright\arraybackslash}p{(\columnwidth - 4\tabcolsep) * \real{0.3333}}@{}}
\toprule
\begin{minipage}[b]{\linewidth}\raggedright
Main-paper claim
\end{minipage} & \begin{minipage}[b]{\linewidth}\raggedright
Supporting analysis
\end{minipage} & \begin{minipage}[b]{\linewidth}\raggedright
Section
\end{minipage} \\
\midrule
Scanner shift is measurable & Scanner decodability, subspace, and
projection analyses & A.1--A.3 \\
Clean rank is unstable & Selection regret, paired bootstraps,
source-target matrix & A.7, A.9, A.15 \\
Exposure must be audited & Controlled exposure experiment & A.5 \\
Class-specific failure modes localize the MLL23 collapse & Per-class
confusion on MLL23 & A.6 \\
Calibration fails off-domain & ECE / adaptive-ECE / NLL / Brier and
temperature scaling & A.13 \\
Label-free selection fails under skew & Marginal entropy and
agreement-style selectors & A.10--A.11 \\
CBR mitigates skewed-prior TTA failure & CBR stress tests, edge cases,
small batches, per-encoder harm counts & A.12 \\
Resolution does not explain the rank flip & DinoBloom 224-vs-518
sensitivity & A.14 \\
Probe-dependence is not a single-head artifact & Linear-head and
local-geometry sensitivity & A.16 \\
\bottomrule
\end{tabular}\end{adjustbox}
\end{table}

Sections A.1--A.6 analyze mechanisms and class-specific failures;
A.7--A.11 analyze model selection; A.12--A.13 analyze adaptation and
calibration; and A.14--A.16 provide resolution, source-target, and
head-sensitivity controls.

\section*{A.1 Scanner identity is linearly decodable from frozen
embeddings}\label{a.1-scanner-identity-is-linearly-decodable-from-frozen-embeddings}

To avoid a class-composition confound (domains differ in WBC prevalence,
so a scanner classifier could exploit class frequency), we build a
sample with \textbf{equal WBC-class composition across domains} (79
images per WBC class per domain, the largest common per-cell count), fit
feature standardization on the \textbf{training split only}, and train
(70/30 split, 5 balanced draws) a 4-way scanner classifier and a 5-way
WBC-class classifier from each frozen encoder's features. We
additionally decode the scanner \textbf{within a single WBC class at a
time} (the tightest control) and report a permutation baseline (shuffled
domain labels).

\begin{table}[ht]\centering\footnotesize\begin{adjustbox}{max width=\textwidth}\begin{tabular}{@{}
  >{\raggedright\arraybackslash}p{(\columnwidth - 6\tabcolsep) * \real{0.2500}}
  >{\raggedright\arraybackslash}p{(\columnwidth - 6\tabcolsep) * \real{0.2500}}
  >{\raggedright\arraybackslash}p{(\columnwidth - 6\tabcolsep) * \real{0.2500}}
  >{\raggedright\arraybackslash}p{(\columnwidth - 6\tabcolsep) * \real{0.2500}}@{}}
\toprule
\begin{minipage}[b]{\linewidth}\raggedright
All 11 encoders (class-balanced)
\end{minipage} & \begin{minipage}[b]{\linewidth}\raggedright
scanner (chance 0.25)
\end{minipage} & \begin{minipage}[b]{\linewidth}\raggedright
WBC5 (chance 0.20)
\end{minipage} & \begin{minipage}[b]{\linewidth}\raggedright
within-class scanner
\end{minipage} \\
\midrule
decodability & 0.990--1.000 (mean 0.997) & 0.782--0.971 (mean 0.891) &
0.974--0.999 (mean 0.990) \\
\bottomrule
\end{tabular}\end{adjustbox}
\end{table}

Even with WBC-class composition equalized across domains and
preprocessing fit only on training data, scanner identity is linearly
decodable with accuracy 0.990--1.000 (mean 0.997), whereas WBC5 class
decodability ranges 0.782--0.971 (mean 0.891); the permutation baseline
sits at chance (mean 0.25). Crucially, scanner remains almost perfectly
decodable \emph{within a single WBC class} (mean 0.990), so this is not
a class-prevalence artifact. In this auxiliary linear-decoding task
acquisition is therefore separable at least as well as biology,
indicating that acquisition information remains strongly represented in
frozen blood-cell embeddings across model families. This extends the
histopathology site-signature / batch-effect finding to single-cell
hematology, across foundation-model families rather than a single
pretraining recipe.

Scanner decodability is saturated across encoders and therefore does not
by itself explain which encoder is most robust (Spearman ρ between
scanner decodability and mean cross-dataset drop ≈ −0.09, not
significant); the remaining variation appears associated with
target-shift magnitude and encoder-specific class geometry (Sec. 5.1). A
source-trained linear probe is fitted in a feature space that retains
strong scanner-associated variation, so under a scanner-associated
domain change the features shift along these directions relative to the
source-fit boundary, contributing to the cross-domain macro-F1 collapse
(Sec. 5). These results are consistent with a substantial first-moment
component (a per-scanner mean offset) to the acquisition shift, which is
what feature re-standardization corrects.

\section*{A.2 Class--scanner subspace
analysis}\label{a.2-classscanner-subspace-analysis}

For each encoder we measure the fraction of the WBC5
class-discriminative subspace energy that lies in the top-3
between-domain (scanner) subspace. Entanglement is small (0.03--0.16)
and does not correlate with the cross-dataset drop (Spearman ρ=−0.03,
p=0.94). Neither saturated scanner decodability (A.1) nor this global
class--scanner overlap explains which encoder is most robust; the
remaining variation appears more closely related to target-shift
magnitude and encoder-specific class geometry (Sec. 5.1) than to a
simple geometric feature-space property.

\section*{A.3 Scanner-direction projection as a diagnostic
intervention}\label{a.3-scanner-direction-projection-as-a-diagnostic-intervention}

We project features onto the orthogonal complement of the top-3
(label-free) between-domain scanner directions, retrain the source
linear probe, and re-evaluate cross-scanner. Projecting away these
directions improves mean cross-target macro-F1 by +0.063 at zero
in-domain cost (in-domain Δ = +0.000). This suggests that source-trained
linear probes partly rely on scanner-associated directions. The effect
is strongest for general encoders (DINOv2-S +0.145, DINOv2-L +0.126,
CLIP-L +0.129) and weak for DinoBloom (+0.01--0.05) and BiomedCLIP (≈0),
indicating that DinoBloom's MLL23 failure is not fully captured by a
removable global scanner-mean axis but reflects its specific learned
class geometry. Scanner-subspace projection is a diagnostic first-moment
intervention, not the deployment method used in the main paper. It
supports the low-order-shift interpretation and complements the CBR
analysis.

\section*{A.4 Effective-robustness residuals are uninformative under
saturated in-domain
accuracy}\label{a.4-effective-robustness-residuals-are-uninformative-under-saturated-in-domain-accuracy}

We examined the confound via Taori effective-robustness residuals (above
or below the in-domain-to-out-of-domain line). Because all frozen FMs
have saturated, near-identical in-domain accuracy (0.98--0.997), this
line is ill-conditioned (no spread on the x-axis) and the residual
re-encodes cross-domain rank rather than a meaningful effective
robustness above the line (DinoBloom-family residual ≈ −0.08 to −0.11
across targets). In this benchmark, the clean-vs-target rank matrix
(Sec. 5.1, main-paper Fig. 1) is more informative than
effective-robustness residuals.

\section*{A.5 Controlled exposure
experiment}\label{a.5-controlled-exposure-experiment}

\textbf{Setup.} We fine-tune a clean encoder (DINOv2-S, no hematology
pretraining) on one target scanner using WBC5 labels, re-extract
features on all four domains, train a fresh Acevedo probe, and measure
the change in target macro-F1 relative to the unfine-tuned encoder.

\textbf{Result.} Fine-tuning on Matek-M8 leaves Acevedo in-domain
unchanged (0.987→0.982, Δ−0.004) but raises the exposed scanner most
(Matek 0.497→0.778, Δ+0.281) versus an unexposed scanner (MLL23
0.614→0.761, Δ+0.148); selective inflation = ΔMatek − ΔMLL23 = +0.133.
Repeating for each scanner gives a diagonal-dominant matrix of macro-F1
gains over the unfine-tuned encoder:

\begin{table}[ht]\centering\footnotesize\begin{adjustbox}{max width=\textwidth}\begin{tabular}{@{}llll@{}}
\toprule
fine-tune ~test & Matek & MLL23 & Raabin \\
\midrule
FT-Matek & +0.317 & +0.189 & +0.325 \\
FT-MLL23 & +0.278 & +0.246 & +0.198 \\
FT-Raabin & +0.101 & +0.061 & +0.321 \\
\bottomrule
\end{tabular}\end{adjustbox}
\end{table}

Fine-tuning produces a diagonal-dominant pattern overall: Matek and
MLL23 gain most from same-scanner exposure, while Raabin is essentially
tied after rounding (+0.325 under FT-Matek vs +0.321 under FT-Raabin).
Subtracting the mean cross-scanner gain still leaves a positive
selective component for all three targets: +0.128 for Matek, +0.121 for
MLL23, and +0.059 for Raabin.

\textbf{Interpretation.} Scanner exposure can selectively inflate
apparent robustness in a controlled supervised setting.

\textbf{Limitation.} This is supervised fine-tuning, not self-supervised
pretraining, so it conflates seeing the scanner with learning the task
(the clean signal is the selective diagonal component). It motivates
exposure reporting and demonstrates the mechanism in principle, but does
not prove DinoBloom-specific leakage inflation; DinoBloom's measured
numbers do not show this inflation (Sec. 5.1).

\section*{A.6 Per-class confusion on
MLL23}\label{a.6-per-class-confusion-on-mll23}

On the MLL23 target, DinoBloom-B mis-predicts lymphocytes as neutrophils
(57\% → neutrophil, only 16\% correct; +26\% → monocyte) and eosinophils
as neutrophils (45\%): minority WBC types collapse into the majority
class off-scanner. DINOv2 retains lymphocytes (recall 0.82, 5-seed mean,
matching the main paper) on the same images, its errors instead leaking
eosinophils toward basophil rather than collapsing minority types into
neutrophils. The Sec. 5.2 DinoBloom lymphocyte collapse is thus
specifically a lymphocyte-to-neutrophil confusion, while DINOv2-B's
robustness reflects a preserved class geometry rather than a global
accuracy offset.

Having localized the failure to acquisition-sensitive class geometry, we
next quantify its consequences for encoder selection.

\section*{A.7 Selection regret under balanced
evaluation}\label{a.7-selection-regret-under-balanced-evaluation}

For each target scanner, regret is the target macro-F1 of the oracle
encoder minus the target macro-F1 of the encoder selected by in-domain
accuracy.

\begin{table}[ht]\centering\footnotesize\begin{adjustbox}{max width=\textwidth}\begin{tabular}{@{}lll@{}}
\toprule
Target shift from Acevedo & Linear regret & 1-NN regret \\
\midrule
Matek-M8 (smaller shift) & 0.000 & 0.000 \\
MLL23-Metafer (largest shift) & 0.151 & 0.088 \\
Raabin (mid shift) & 0.064 & 0.129 \\
\bottomrule
\end{tabular}\end{adjustbox}
\end{table}

Selecting by in-domain accuracy costs up to about 15 macro-F1 points on
the most-shifted target (one would pick DinoBloom-L; RedDino is far
better) and is near-free on the closest target. Clean accuracy is thus
an unreliable deployment selector exactly where deployment risk is
highest. This regret is computed on balanced targets; under realistic
class imbalance the absolute numbers shift, and label-free alternatives
to clean-accuracy selection also fail (A.10), so we recommend selecting
encoders using representative cross-domain validation.

\section*{A.8 Generality across FM
families}\label{a.8-generality-across-fm-families}

RedDino (MICCAI 2025; an RBC-focused DINOv2, described as RBC-focused
rather than never trained on white cells) is the most robust encoder on
the WBC MLL23 target (0.704, highest of 15), while the WBC-specialist
DinoBloom-L falls to 10th of 15 (0.552) and the two pathology FMs
diverge sharply (Lunit-DINO 0.609 vs Phikon 0.410). Specialization
therefore does not guarantee robustness. This is a second-FM
confirmation of the rank-stability issue, not an
in-vs-out-of-pretraining contrast: RedDino, like DinoBloom, was trained
on overlapping public hematology data.

\section*{A.9 Paired-bootstrap stability of the headline MLL23 rank
differences}\label{a.9-paired-bootstrap-stability-of-the-headline-mll23-rank-differences}

We assess whether the MLL23 rank differences are statistically stable.
We compute a paired bootstrap of the macro-F1 difference for the four
load-bearing comparisons: resample the 3,816 MLL23 test images with
replacement (B=4,000), recompute each encoder's macro-F1 on the same
resampled set (paired, removing shared example-level variance), average
over the 5 source-split seeds, and take the difference. Point estimates
reproduce the main-paper Table 1.

\begin{table}[ht]\centering\footnotesize\begin{adjustbox}{max width=\textwidth}\begin{tabular}{@{}
  >{\raggedright\arraybackslash}p{(\columnwidth - 6\tabcolsep) * \real{0.2500}}
  >{\raggedright\arraybackslash}p{(\columnwidth - 6\tabcolsep) * \real{0.2500}}
  >{\raggedright\arraybackslash}p{(\columnwidth - 6\tabcolsep) * \real{0.2500}}
  >{\raggedright\arraybackslash}p{(\columnwidth - 6\tabcolsep) * \real{0.2500}}@{}}
\toprule
\begin{minipage}[b]{\linewidth}\raggedright
Comparison (MLL23)
\end{minipage} & \begin{minipage}[b]{\linewidth}\raggedright
Δ macro-F1
\end{minipage} & \begin{minipage}[b]{\linewidth}\raggedright
95\% CI
\end{minipage} & \begin{minipage}[b]{\linewidth}\raggedright
P(Δ\textgreater0)
\end{minipage} \\
\midrule
RedDino − DinoBloom-L (in-domain \#1) & +0.151 & {[}+0.134, +0.168{]} &
1.000 \\
RedDino − DinoBloom-S & +0.033 & {[}+0.017, +0.049{]} & 1.000 \\
DinoBloom-S − DinoBloom-L & +0.119 & {[}+0.105, +0.133{]} & 1.000 \\
Lunit-DINO − Phikon & +0.200 & {[}+0.184, +0.215{]} & 1.000 \\
\bottomrule
\end{tabular}\end{adjustbox}
\end{table}

All four CIs exclude zero. The in-domain best, DinoBloom-L, is dethroned
on MLL23: both RedDino and the smaller DinoBloom-S significantly
outperform it. The pathology-FM divergence, with Lunit-DINO well above
Phikon, is also statistically stable. The paired design yields tighter,
more appropriate intervals than per-encoder CI overlap because the
encoders are scored on identical examples.

\FloatBarrier
\section*{A.10 Marginal prediction entropy fails under realistic
class-prior
shift}

\begin{figure}[t]\centering
\includegraphics[width=\textwidth]{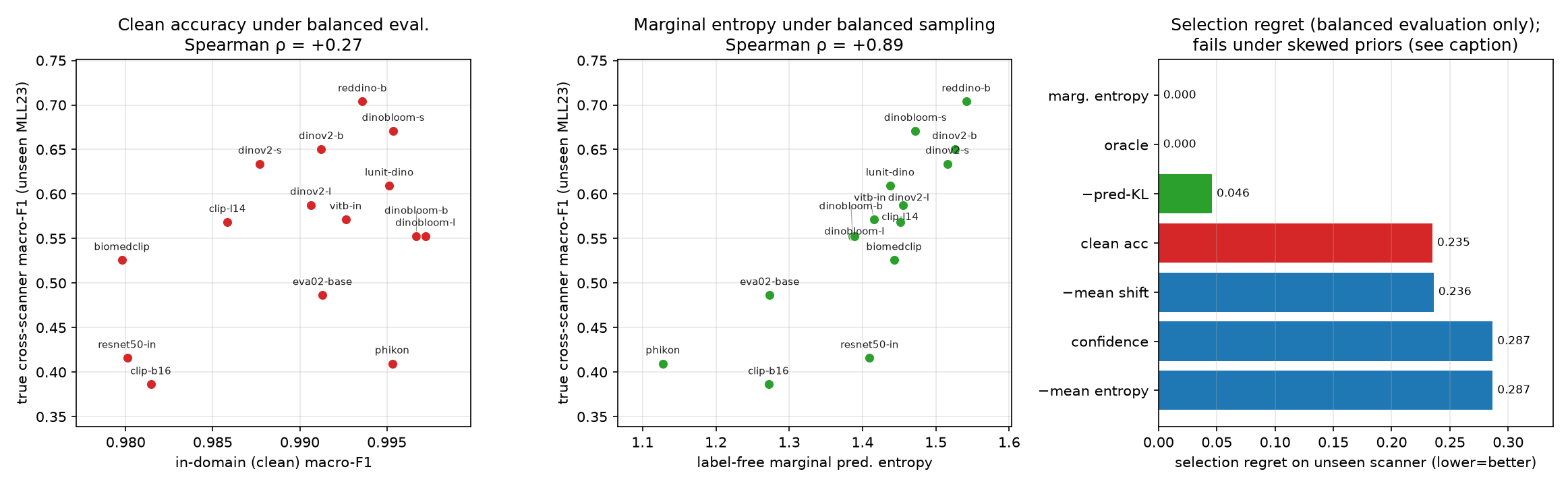}
\caption{(A.10) Marginal prediction entropy appears oracle-matching under \emph{balanced} target sampling but
fails under realistic skewed WBC priors. We report this as a failure case, not a recommended selection
rule.}\label{fig:appD}
\end{figure}
\label{a.10-marginal-prediction-entropy-fails-under-realistic-class-prior-shift}

Since clean accuracy mis-ranks encoders (Sec. 5.1), we assess whether a
statistic from the source probe on unlabeled target images can rank them
instead. We score five label-free signals (Spearman vs true target
macro-F1 across encoders, per source-to-target): \textbf{conf} (mean
max-softmax); \textbf{neg\_ent} (−mean per-sample entropy);
\textbf{neg\_shift} (−RMS first-moment shift); \textbf{neg\_predKL} (−KL
of the predicted-label distribution from the source prior); and
\textbf{marg\_ent}, the entropy of the marginal prediction
\(H(\frac1N\sum_i p_i)\), a prediction-diversity / collapse detector.
The figure in this section shows the balanced-evaluation result. The
regret here uses the label-free selector benchmark: each
unlabeled-target statistic ranks the 15 encoders and is scored against
the true target macro-F1, averaged across the source-to-target
evaluations. This differs from A.7, which reports Acevedo-source
in-domain-accuracy selection regret on balanced targets.

Under balanced target sampling, marginal prediction entropy appears to
select the oracle encoder:

\begin{table}[ht]\centering\footnotesize\begin{adjustbox}{max width=\textwidth}\begin{tabular}{@{}
  >{\raggedright\arraybackslash}p{(\columnwidth - 6\tabcolsep) * \real{0.2500}}
  >{\raggedright\arraybackslash}p{(\columnwidth - 6\tabcolsep) * \real{0.2500}}
  >{\raggedright\arraybackslash}p{(\columnwidth - 6\tabcolsep) * \real{0.2500}}
  >{\raggedright\arraybackslash}p{(\columnwidth - 6\tabcolsep) * \real{0.2500}}@{}}
\toprule
\begin{minipage}[b]{\linewidth}\raggedright
predictor
\end{minipage} & \begin{minipage}[b]{\linewidth}\raggedright
mean ρ, MLL23 (balanced)
\end{minipage} & \begin{minipage}[b]{\linewidth}\raggedright
mean ρ, other targets
\end{minipage} & \begin{minipage}[b]{\linewidth}\raggedright
selector-benchmark regret (balanced)
\end{minipage} \\
\midrule
clean accuracy & +0.27 & +0.51 & 0.235 \\
conf / neg\_ent & +0.04 / +0.00 & −0.16 / −0.23 & 0.287 \\
neg\_shift & +0.30 & +0.59 & 0.236 \\
neg\_predKL & +0.50 & +0.64 & 0.046 \\
marg\_ent & +0.89 & +0.59 & 0.000 (= oracle) \\
\bottomrule
\end{tabular}\end{adjustbox}
\end{table}

This is a balanced-sampling artifact. Resampling the MLL23 target to
realistic class priors:

\begin{table}[ht]\centering\footnotesize\begin{adjustbox}{max width=\textwidth}\begin{tabular}{@{}lll@{}}
\toprule
target prior & marg\_ent regret & clean-accuracy regret \\
\midrule
balanced & 0.002 & 0.162 \\
peripheral-blood-like & 0.290 & 0.151 \\
neutrophil-heavy & 0.372 & 0.139 \\
lymphocyte-poor & 0.252 & 0.122 \\
\bottomrule
\end{tabular}\end{adjustbox}
\end{table}

Under every evaluated skewed prior, marg\_ent's selection regret rises
to 0.25--0.37 and clean accuracy beats it; the regret remains elevated
across K=8--400, so it is not a sample-size effect. Under label shift a
robust encoder should produce a skewed (low-entropy) marginal, so
rewarding prediction diversity penalizes the right encoder. We therefore
treat marginal entropy as a failure case, not a recommended selector.

\section*{A.11 Comparison with agreement-style label-free
selectors}\label{a.11-comparison-with-agreement-style-label-free-selectors}

Agreement-on-the-Line methods (formally cited in the main paper) predict
out-of-distribution behavior from the agreement between models on
unlabeled target data. We compare marginal prediction entropy with two
agreement-style selectors: \textbf{consensus} (mean pairwise
prediction-agreement of an encoder with all others on the unlabeled
target) and \textbf{agl\_acc} (an AgL-style accuracy estimate, clean
accuracy scaled by the encoder's OOD/ID agreement ratio).

\begin{table}[ht]\centering\footnotesize\begin{adjustbox}{max width=\textwidth}\begin{tabular}{@{}
  >{\raggedright\arraybackslash}p{(\columnwidth - 6\tabcolsep) * \real{0.2500}}
  >{\raggedright\arraybackslash}p{(\columnwidth - 6\tabcolsep) * \real{0.2500}}
  >{\raggedright\arraybackslash}p{(\columnwidth - 6\tabcolsep) * \real{0.2500}}
  >{\raggedright\arraybackslash}p{(\columnwidth - 6\tabcolsep) * \real{0.2500}}@{}}
\toprule
\begin{minipage}[b]{\linewidth}\raggedright
selector
\end{minipage} & \begin{minipage}[b]{\linewidth}\raggedright
selector-benchmark regret (balanced)
\end{minipage} & \begin{minipage}[b]{\linewidth}\raggedright
mean ρ vs true F1
\end{minipage} & \begin{minipage}[b]{\linewidth}\raggedright
needs
\end{minipage} \\
\midrule
oracle & 0.000 & -- & target labels \\
marg\_ent & 0.000 & +0.89 & one encoder, unlabeled target \\
agl\_acc (AgL-style) & 0.051 & +0.88 & all encoders' predictions + clean
accuracy \\
consensus agreement & 0.051 & +0.88 & all encoders' predictions \\
clean accuracy & 0.235 & +0.27 & source labels only \\
\bottomrule
\end{tabular}\end{adjustbox}
\end{table}

Under balanced sampling, agreement-style selectors outperform clean
accuracy. Because they were not evaluated under skewed priors here,
their robustness to clinical WBC class-prior shift remains unestablished
(they reward broad class usage and so may share the class-balance
vulnerability of marginal entropy, but we do not measure this). Marginal
prediction entropy, which was directly stress-tested under skew (A.10),
fails consistently. We therefore do not identify a label-free selector
with demonstrated robustness to the evaluated class-prior shifts; robust
label-free encoder selection under label shift remains open.

The same class-prior assumption that destabilizes model selection also
affects target-statistics adaptation, motivating the class-balanced
analysis below.

\section*{A.12 CBR under label-shifted target
priors}\label{a.12-cbr-under-label-shifted-target-priors}

\subsection*{A.12.1 Main skewed-prior
results}\label{a.12.1-main-skewed-prior-results}

This is \textbf{one canonical experiment} that drives both Figure 3 and
every CBR headline number: source = Acevedo, all 15 encoders, 5 source
seeds, 3 targets, 6 target priors (balanced, Dirichlet-A,
lymphocyte-poor, Dirichlet-B, peripheral-blood-like, neutrophil-heavy),
25 target draws per cell, fixed target batch N=400, primary metric
fixed-WBC5 macro-F1. We measure macro-F1 gain over the no-adaptation
baseline; the 18 scenarios are the 3 targets × 6 priors. CIs are a
hierarchical bootstrap over scenarios (A.12.6).

\begin{table}[ht]\centering\footnotesize\begin{adjustbox}{max width=\textwidth}\begin{tabular}{@{}llll@{}}
\toprule
method & mean {[}95\% CI{]} & worst scenario & hurts \\
\midrule
tgtstd & +0.032 {[}−0.006, +0.070{]} & −0.112 & 6/18 \\
SHOT/IM & −0.029 {[}−0.087, +0.031{]} & −0.265 & 10/18 \\
CBR & +0.059 {[}+0.046, +0.073{]} & +0.007 & 0/18 \\
CBR-mean & +0.053 {[}+0.038, +0.065{]} & +0.002 & 0/18 \\
CBR-oracle & +0.098 {[}+0.081, +0.113{]} & +0.023 & 0/18 \\
\bottomrule
\end{tabular}\end{adjustbox}
\end{table}

Here tgtstd = global target feature standardization, CBR-mean = balanced
mean only, and CBR-oracle = true-label-balanced statistics (a
non-deployable upper bound). Across the 18 target-prior scenarios, CBR
produced a positive mean gain in every evaluated scenario (mean +0.059,
median +0.057, range {[}+0.007, +0.109{]}, hurt 0/18), whereas global
target standardization helps on balanced/mild priors but hurts in 6/18
(it falls to −0.07/−0.08 on the clinical and neutrophil-heavy priors),
and SHOT/IM hurts in 10/18 (its diversity regularizer drives predictions
toward a uniform marginal, which is wrong under skew). CBR uses
pseudo-label-balanced, class-prior-reduced statistics: the balanced mean
\(\bar\mu=\frac1{|\mathcal C|}\sum_c\mathrm{mean}(X_t[\hat c{=}c])\) and
a pooled within-class standard deviation. An ablation shows the balanced
first-moment term drives the gain (+0.053 alone), tying CBR to the
first-moment mechanism (A.1, A.3); the balanced within-class standard
deviation adds little. CBR recovers about 61\% of the true-label oracle,
so pseudo-label noise costs something but CBR remains positive. CBR is
label-free, training-free, and single-batch; it is transductive,
pseudo-label-dependent, and evaluated on 5-class WBC. Decomposed per
encoder (15 encoders × 18 scenarios = 270 cells, each cell a mean over 5
seeds × 25 draws), the mean CBR gain is positive for all 15 encoders,
but 29/270 cells are negative: the two most skewed priors account for 23
of the 29 and the DinoBloom family for 15 (worst cell −0.090,
DinoBloom-S, Matek/peripheral-blood-like; DinoBloom mean gains are
smallest, +0.002 to +0.018, consistent with their pretraining exposure
to the targets, Sec.~5.1). Under the same decomposition tgtstd is
negative in 96/270 cells (worst −0.24) and SHOT/IM in 149/270 (worst
−0.37), each harming every encoder in at least one scenario. The
headline ``positive in all 18 scenarios'' therefore holds at the
scenario-mean level, not uniformly per encoder.

\subsection*{A.12.2 Pseudo-class edge
cases}\label{a.12.2-pseudo-class-edge-cases}

A pseudo-class predicted for no target cell in the batch is omitted from
the balanced mean and standard deviation; a singleton pseudo-class
contributes to the balanced mean but not the within-class variance
(undefined for \(n{=}1\)); if the source probe collapses all target
cells onto a single pseudo-class, CBR falls back to the source standard
deviation. These rules keep CBR well-defined under the skewed and small
batches that destabilize global target statistics, and are exercised by
the small-batch stress test (A.12.5).

\subsection*{A.12.3 Target-prior
definitions}\label{a.12.3-target-prior-definitions}

\begin{table}[ht]\centering\footnotesize\begin{adjustbox}{max width=\textwidth}\begin{tabular}{@{}llllll@{}}
\toprule
Prior & Neu & Lym & Mon & Eos & Bas \\
\midrule
Balanced & 0.20 & 0.20 & 0.20 & 0.20 & 0.20 \\
Peripheral-blood-like & 0.62 & 0.30 & 0.05 & 0.02 & 0.01 \\
Neutrophil-heavy & 0.80 & 0.05 & 0.05 & 0.05 & 0.05 \\
Lymphocyte-poor & 0.40 & 0.02 & 0.19 & 0.19 & 0.20 \\
Dirichlet-A & 0.45 & 0.10 & 0.25 & 0.15 & 0.05 \\
Dirichlet-B & 0.10 & 0.50 & 0.15 & 0.20 & 0.05 \\
\bottomrule
\end{tabular}\end{adjustbox}
\end{table}

Class order is neutrophil, lymphocyte, monocyte, eosinophil, basophil.
The peripheral-blood-like prior approximates a normal adult WBC
differential; the neutrophil-heavy prior stress-tests inflammatory or
infection-like deployment batches (illustrative, not a clinical
reference range). Each prior is applied by resampling the target test
set to those class frequencies; the no-adaptation baseline uses the same
resampled set, so the reported gain isolates the adaptation, not the
resampling. Priors are applied by sampling without replacement (which
can cap a rare class when the target pool is small); the realized class
proportions for each draw are recorded in the canonical results
file.

\subsection*{A.12.4 Comparison with BBSE label-shift
correction}\label{a.12.4-comparison-with-bbse-label-shift-correction}

BBSE (Lipton et al., cited in the main paper) estimates target/source
importance weights \(w_y\) by solving \(\hat C w = \hat\mu\) (\(\hat C\)
= source confusion, \(\hat\mu\) = target predicted marginal), then
reweights posteriors. Over the same 18 canonical target-prior scenarios
(A.12.1):

\begin{table}[ht]\centering\footnotesize\begin{adjustbox}{max width=\textwidth}\begin{tabular}{@{}llll@{}}
\toprule
method & mean gain & worst & hurts \\
\midrule
CBR & +0.059 & +0.007 & 0/18 \\
BBSE & −0.035 & −0.051 & 18/18 \\
CBR + BBSE & +0.055 & +0.015 & 0/18 \\
\bottomrule
\end{tabular}\end{adjustbox}
\end{table}

BBSE hurts macro-F1 in all 18 scenarios: it corrects the label prior
(reweighting toward the estimated majority class), which can help
balanced accuracy but hurts minority-class F1, and its prior estimate
from a scanner-shifted, partly collapsing probe is unreliable. This
indicates the failure is not a pure label-prior shift that
prediction-space BBSE can solve; scanner-associated feature shift
remains load-bearing. CBR+BBSE ≈ CBR, so label-shift correction is not
the missing ingredient.

\subsection*{A.12.5 Small-batch and missing-pseudo-class stress
test}\label{a.12.5-small-batch-and-missing-pseudo-class-stress-test}

The most demanding regime for CBR is tiny transductive target batches
under realistic skew, where rare pseudo-classes are often absent. We
sample K unlabeled target images at a skewed prior, adapt using those K
(CBR vs vanilla tgtstd), and predict them; 15 encoders × 3 seeds × 20
draws.

\begin{table}[ht]\centering\footnotesize\begin{adjustbox}{max width=\textwidth}\begin{tabular}{@{}lllllll@{}}
\toprule
K & prior & base F1 & tgtstd Δ & CBR Δ & CBR hurt \% & \#pseudo (of
5) \\
\midrule
16 & PB-like & 0.544 & −0.003 & +0.024 & 46.5 & 3.0 \\
16 & N-heavy & 0.424 & −0.011 & +0.072 & 25.8 & 3.4 \\
16 & L-poor & 0.470 & +0.144 & +0.092 & 19.9 & 3.8 \\
32 & PB-like & 0.472 & −0.017 & +0.040 & 34.1 & 3.6 \\
32 & N-heavy & 0.442 & −0.027 & +0.069 & 25.8 & 3.9 \\
32 & L-poor & 0.435 & +0.122 & +0.085 & 15.3 & 4.3 \\
64 & PB-like & 0.430 & −0.057 & +0.015 & 37.4 & 4.1 \\
64 & N-heavy & 0.434 & −0.065 & +0.059 & 24.3 & 4.2 \\
64 & L-poor & 0.422 & +0.112 & +0.081 & 12.8 & 4.5 \\
128 & PB-like & 0.428 & −0.062 & +0.020 & 33.3 & 4.4 \\
128 & N-heavy & 0.444 & −0.078 & +0.057 & 21.2 & 4.5 \\
128 & L-poor & 0.434 & +0.113 & +0.082 & 8.7 & 4.7 \\
\bottomrule
\end{tabular}\end{adjustbox}
\end{table}

PB-like = peripheral-blood-like; N-heavy = neutrophil-heavy; L-poor =
lymphocyte-poor. Aggregated over the 12 K-prior cells, vanilla tgtstd
has mean +0.014 (worst −0.078, hurts in 8 of 12 cells) while CBR has
mean +0.058 (worst +0.015, positive in every cell): CBR is positive on
average in every K-prior cell, even at K=16 with absent rare classes.
The per-draw harm rate is non-trivial on the smallest batches under
strong skew (up to about 46\% at K=16) even though the mean is positive,
so CBR is reliable in expectation but high-variance on tiny imbalanced
batches, improving with K. We recommend K≥32. In practice, these results
support using CBR with batches of at least roughly 32 unlabeled target
cells, or aggregating statistics over multiple small batches when a
deployment batch is extremely imbalanced.

\subsection*{A.12.6 Bootstrap
interpretation}\label{a.12.6-bootstrap-interpretation}

The CIs reported for CBR are a \textbf{hierarchical bootstrap clustered
by scenario}: encoder×seed×draw observations within a target-prior
scenario are not independent, so we first reduce each of the 18
scenarios to its mean gain, then resample the 18 scenario means with
replacement (2,000 times), take the mean each time, and report the
2.5/97.5 percentiles. This respects the dependence structure rather than
treating every encoder×seed×draw row as an independent replicate; the
scenario range and hurt count are reported alongside the CI.

\section*{A.13 Calibration metrics and temperature-scaling
transfer}\label{a.13-calibration-metrics-and-temperature-scaling-transfer}

Source-trained linear probe (source = Acevedo, 5 seeds), 15 encoders.
Metrics: top-label ECE (15 equal-width bins), adaptive-ECE (15
equal-mass bins), NLL, and multiclass Brier. Temperature scaling fits T
on the held-out source test split (minimizing NLL) and applies it to the
target without refitting (the deployable case); oracle-T fits T on the
target (using target labels, an upper bound, not deployable). The CBR
arm re-standardizes target features before scoring.

\begin{table}[ht]\centering\footnotesize\begin{adjustbox}{max width=\textwidth}\begin{tabular}{@{}
  >{\raggedright\arraybackslash}p{(\columnwidth - 12\tabcolsep) * \real{0.1429}}
  >{\raggedright\arraybackslash}p{(\columnwidth - 12\tabcolsep) * \real{0.1429}}
  >{\raggedright\arraybackslash}p{(\columnwidth - 12\tabcolsep) * \real{0.1429}}
  >{\raggedright\arraybackslash}p{(\columnwidth - 12\tabcolsep) * \real{0.1429}}
  >{\raggedright\arraybackslash}p{(\columnwidth - 12\tabcolsep) * \real{0.1429}}
  >{\raggedright\arraybackslash}p{(\columnwidth - 12\tabcolsep) * \real{0.1429}}
  >{\raggedright\arraybackslash}p{(\columnwidth - 12\tabcolsep) * \real{0.1429}}@{}}
\toprule
\begin{minipage}[b]{\linewidth}\raggedright
metric
\end{minipage} & \begin{minipage}[b]{\linewidth}\raggedright
source (uncal.)
\end{minipage} & \begin{minipage}[b]{\linewidth}\raggedright
target (uncal.)
\end{minipage} & \begin{minipage}[b]{\linewidth}\raggedright
target + source-T
\end{minipage} & \begin{minipage}[b]{\linewidth}\raggedright
target + oracle-T
\end{minipage} & \begin{minipage}[b]{\linewidth}\raggedright
target + CBR
\end{minipage} & \begin{minipage}[b]{\linewidth}\raggedright
target + CBR + source-T
\end{minipage} \\
\midrule
ECE & 0.004 & 0.348 & 0.315 & 0.070 & 0.290 & 0.254 \\
adaptive-ECE & 0.003 & 0.348 & 0.315 & 0.073 & 0.290 & 0.254 \\
NLL & 0.031 & 3.200 & 2.515 & 1.205 & 2.213 & 1.772 \\
Brier & 0.013 & 0.783 & 0.750 & 0.597 & 0.683 & 0.652 \\
\bottomrule
\end{tabular}\end{adjustbox}
\end{table}

Calibration collapses off-domain (ECE 0.004→0.348, about 80×; NLL
0.031→3.20; Brier 0.013→0.78); the reliability diagram (main-paper Fig.
2) shows target predictions below the diagonal (confidently wrong).
Source-domain temperature scaling is insufficient for deployment on a
shifted scanner: applying source-fitted T barely moves target ECE
(0.348→0.315), because the source probe is already near-perfectly
calibrated (fitted T≈1). Only oracle target temperature scaling
substantially improves it (ECE→0.070, still above the in-domain 0.004:
it improves but does not restore), and it requires target labels. CBR
partially improves calibration (CBR alone 0.348→0.290 ECE; CBR+source-T
0.254 ECE, −27\%; NLL −45\%), but residual ECE (0.254) remains far above
the in-domain 0.004. Robustness (Axis A) and calibration (Axis B) are
distinct deployment failures that must both be evaluated per
scanner/site.

\textbf{Per-encoder, per-target ECE.} Decomposed per encoder and target
(15 × 3 cells), CBR+source-T lowers ECE in 43 of the 45 encoder×target
cells; the two exceptions are Phikon and RedDino on Raabin (ECE
increases by $\approx$0.03), and CBR alone helps in 40/45 (additionally
failing for Lunit-DINO on MLL23/Raabin and ViT-B on Matek).
Miscalibration is also uneven across targets: 13 of 15 encoders have
their worst uncalibrated ECE on Raabin (0.31--0.63), so ``partially
improves calibration'' holds for most, not all, encoder×target pairs,
and residual miscalibration is largest exactly where calibration was
worst.

\textbf{Few-shot recalibration.} Source-T does not transfer and oracle-T
needs all target labels, but a small labeled per-scanner calibration set
suffices. Fitting T on K labeled target images and applying it to the
rest (15 encoders × 3 targets):

\begin{table}[ht]\centering\footnotesize\begin{adjustbox}{max width=\textwidth}\begin{tabular}{@{}ll@{}}
\toprule
arm & target ECE \\
\midrule
uncalibrated & 0.351 \\
+ source-T (no transfer) & 0.308 \\
+ few-shot target-T, K=4 & 0.165 \\
+ few-shot target-T, K=8 & 0.124 \\
+ few-shot target-T, K=16 & 0.100 \\
+ few-shot target-T, K=32 & 0.088 \\
+ few-shot target-T, K=64 & 0.081 \\
+ oracle-T (all target labels) & 0.074 \\
\bottomrule
\end{tabular}\end{adjustbox}
\end{table}

About 16--32 labeled target images recover most of the calibration
(within +0.02 of the oracle at K=32), turning Axis B into an actionable
per-scanner protocol.

We finally test whether the conclusions depend on input resolution,
source domain, or the operational head.

\section*{A.14 DinoBloom resolution sensitivity (224 vs
518)}\label{a.14-dinobloom-resolution-sensitivity-224-vs-518}

The DinoBloom checkpoint name encodes a 224 input size, but its
configuration inherits a 518 default from DINOv2, so running it at 224
(our benchmark's common resolution) could handicap it. On a controlled
same-subset comparison (150 images per class, Acevedo source to MLL23
target), extracting each DinoBloom variant at both 224 and 518:

\begin{table}[ht]\centering\footnotesize\begin{adjustbox}{max width=\textwidth}\begin{tabular}{@{}llll@{}}
\toprule
variant & MLL23 macro-F1 @224 & @518 & Δ(518−224) \\
\midrule
DinoBloom-S & 0.643 & 0.698 & +0.054 \\
DinoBloom-B & 0.588 & 0.611 & +0.023 \\
DinoBloom-L & 0.559 & 0.628 & +0.068 \\
\bottomrule
\end{tabular}\end{adjustbox}
\end{table}

Running at 518 helps DinoBloom (+0.02--0.07, most for the large model),
so 224 mildly handicaps it. The qualitative conclusion is
resolution-robust: at 518 the in-domain-best DinoBloom-L (0.628) is
still beaten by the smaller DinoBloom-S (0.698). Giving every encoder
its 518 input, DinoBloom-L (0.628) \textless{} RedDino (0.646)
\textless{} DinoBloom-S (0.698), so the in-domain-best DinoBloom-L is
still beaten by both a smaller DinoBloom and the RBC-focused RedDino.
The exact rank and magnitude are resolution-dependent (DinoBloom-L's
10th-of-15 at 224 would improve to a middle rank at 518); we present the
224 benchmark as the common-resolution setting, while the reliability
message (in-domain best is not most robust, and specialization and size
do not guarantee robustness) holds at both resolutions.

\section*{A.15 Full source--target rank matrix (both
probes)}\label{a.15-full-sourcetarget-rank-matrix-both-probes}

Axis A's headline uses Acevedo as source. To check it is not an Acevedo
artifact, we run the full source-target matrix (each domain as source)
for \textbf{both probes}. We report Spearman ρ(in-domain rank, target
rank) across the 15 encoders, and (for the linear probe) whether the
in-domain-best encoder stays first on the target.

\textbf{Linear probe.}

\begin{table}[ht]\centering\footnotesize\begin{adjustbox}{max width=\textwidth}\begin{tabular}{@{}lllll@{}}
\toprule
source ↓ / target → & Matek & MLL23 & Raabin & Acevedo \\
\midrule
Acevedo & 0.62 (no flip) & 0.27 (flip) & 0.43 (flip) & -- \\
Matek & -- & 0.24 (flip) & 0.52 (flip) & 0.68 (no flip) \\
MLL23 & 0.46 (flip) & -- & 0.43 (no flip) & 0.20 (flip) \\
Raabin & 0.67 (no flip) & 0.31 (flip) & -- & 0.53 (flip) \\
\bottomrule
\end{tabular}\end{adjustbox}
\end{table}

Across all 12 (source, target) cells, ρ ∈ {[}0.20, 0.68{]}, median 0.45
(mean 0.45), and the in-domain-best encoder is dethroned on the target
in 8/12 cells. Clean linear-probe accuracy is a poor-to-moderate
predictor of cross-dataset robustness from every source, and
MLL23-as-target is the hardest (ρ 0.24--0.31) from every source,
consistent with it being the largest distribution shift. When Acevedo is
the target rather than the source, DinoBloom transfers best to it (rank
1), consistent with Acevedo being DinoBloom's held-out, canonical,
high-SNR domain rather than evidence of leakage (Sec. 5.1).

\textbf{1-NN probe.} Repeating the full matrix with the 1-NN probe:

\begin{table}[ht]\centering\footnotesize\begin{adjustbox}{max width=\textwidth}\begin{tabular}{@{}lllll@{}}
\toprule
source ↓ / target → & Matek & MLL23 & Raabin & Acevedo \\
\midrule
Acevedo & 0.80 & 0.77 & 0.34 & -- \\
Matek & -- & 0.88 & 0.61 & 0.86 \\
MLL23 & 0.77 & -- & 0.65 & 0.66 \\
Raabin & 0.58 & 0.56 & -- & 0.62 \\
\bottomrule
\end{tabular}\end{adjustbox}
\end{table}

Across the 12 1-NN cells, ρ ∈ {[}0.34, 0.88{]}, median 0.65 (mean 0.67).

\textbf{Probe-dependence.} 1-NN rank transfer is higher on average than
the linear probe (median 0.65 vs 0.45), so local nearest-neighbor
geometry transfers more consistently than a source-fitted linear
boundary. It is not universally predictive, however: Acevedo→Raabin
drops to ρ=0.34 under 1-NN (below the corresponding linear ρ=0.43), and
several 1-NN cells remain in the 0.55--0.65 range. Robustness rankings
are therefore probe-dependent, and neither clean-domain probe is a
universally reliable selector; MLL23-as-target remains among the hardest
for both probes. The benchmark thus evaluates encoders together with the
probe used to operationalize them, not a representation-only property.

\section*{A.16 The probe-dependence is not a single-head
artifact}\label{a.16-the-probe-dependence-is-not-a-single-head-artifact}

To verify that the linear mis-ranking is not an artifact of the specific
linear head (logistic regression, C=1), we re-evaluate source→target
rank transfer (Acevedo source, Spearman ρ across the 15 encoders) for a
range of \textbf{source-only} heads (no target tuning): logistic
regression at C ∈ \{1e-4, 1e-2, 1, 1e2\}, a linear SVM, a
nearest-centroid classifier, a cosine nearest-centroid, and 1-NN. Every
head uses the exact protocol of the main benchmark and A.15: the full
Acevedo source, the same five shared source train/test splits,
source-fitted standardization, and macro-F1 averaged over seeds. The
logistic C=1 and 1-NN rows are therefore \emph{by construction} the same
experiment as the Acevedo row of A.15 and agree with it exactly.

\begin{table}[ht]\centering\footnotesize\begin{adjustbox}{max width=\textwidth}\begin{tabular}{@{}llllll@{}}
\toprule
head & in-domain F1 & ρ→Matek & ρ→MLL23 & ρ→Raabin & mean ρ \\
\midrule
logistic C=1e-4 & 0.967 & 0.40 & 0.22 & 0.33 & 0.32 \\
logistic C=1e-2 & 0.991 & 0.47 & 0.13 & 0.28 & 0.29 \\
logistic C=1 & 0.990 & 0.62 & 0.27 & 0.43 & 0.44 \\
logistic C=1e2 & 0.989 & 0.70 & 0.38 & 0.55 & 0.54 \\
linear SVM & 0.990 & 0.58 & 0.36 & 0.00 & 0.32 \\
nearest-centroid & 0.890 & 0.45 & 0.61 & 0.24 & 0.43 \\
cosine centroid & 0.892 & 0.58 & 0.69 & 0.09 & 0.45 \\
1-NN & 0.943 & 0.80 & 0.77 & 0.34 & 0.64 \\
\bottomrule
\end{tabular}\end{adjustbox}
\end{table}

Every linear head reaches high in-domain macro-F1 (0.97--0.99) yet
transfers weakly to the most-shifted target (MLL23 ρ 0.13--0.38), so the
rank instability is not specific to the C=1 logistic probe: it persists
across the full tested regularization range (C from 1e-4 to 1e2, six
orders of magnitude) and under a linear SVM. The local-geometry heads
(nearest-centroid, cosine, 1-NN) are more stable on MLL23 (ρ 0.61--0.77)
at an in-domain cost (0.89--0.94). No head is universally reliable,
however: on Raabin the linear SVM and cosine centroid fall to ρ
0.00/0.09, and even the best head on average (1-NN, mean ρ 0.64) is far
from a safe selector. This confirms that clean-domain \emph{linear}
selection is unstable across reasonable head choices on the most-shifted
target, while local neighbor geometry transfers more consistently on
average; robustness rankings depend on the operationalizing probe, not
the representation alone.

\section*{Synthesis}\label{synthesis}

Together, these controls support the main conclusion: favorable source
accuracy and confidence do not guarantee target reliability, and the
operational probe, the target class prior, and the acquisition domain
all influence deployment performance.

\end{document}